\documentclass{article}

\PassOptionsToPackage{numbers, compress}{natbib}

\usepackage[final]{neurips_2022}

\usepackage[utf8]{inputenc} 
\usepackage[T1]{fontenc}    
\usepackage{hyperref}       
\usepackage{url}            
\usepackage{booktabs}       
\usepackage{amsfonts}       
\usepackage{nicefrac}       
\usepackage{microtype}      
\usepackage{xcolor}         
\usepackage{amsmath}
\usepackage{graphicx}
\usepackage{wrapfig}
\usepackage{multirow}
\usepackage{makecell}
\usepackage{caption}
\usepackage{tikz}
\usepackage[resetlabels]{multibib}
\newcites{new}{References}
\usepackage[symbol*]{footmisc}
\usepackage{verbatim}
\usepackage{perpage} 
\MakePerPage{footnote} 

\newcommand\tikzmark[1]{\tikz[remember picture] \node (#1) {};}

\newcommand\CoAuthorMark{\footnotemark[\arabic{footnote}]}

\title{Separable PINN:\\ Mitigating the Curse of Dimensionality in Physics-Informed Neural Networks}

\author{
    Junwoo Cho$^{1}$\thanks{Equal contribution, alphabetically ordered.}\quad Seungtae Nam$^{1}$\protect\CoAuthorMark\quad Hyunmo Yang$^{1}$\quad \textbf{Seok-Bae Yun}$^{2}$\quad \textbf{Youngjoon Hong}$^{2}$\\ \textbf{Eunbyung Park}$^{1, 3}$\thanks{Corresponding author.}\\[3pt]
    $^1$Department of Artificial Intelligence\quad $^2$Department of Mathematics\\ $^3$Department of Electrical and Computer Engineering\\[3pt]
    Sungkyunkwan University\\[3pt]
    \texttt{\{jwcho000, stnamjef, cms8033, sbyun01, hongyj, epark\}@skku.edu}
}

\begin{document}

\maketitle

\vspace{-1em}

\begin{abstract}
Physics-informed neural networks (PINNs) have emerged as new data-driven PDE solvers for both forward and inverse problems.
While promising, the expensive computational costs to obtain solutions often restrict their broader applicability.
We demonstrate that the computations in automatic differentiation (AD) can be significantly reduced by leveraging forward-mode AD when training PINN.
However, a naive application of forward-mode AD to conventional PINNs results in higher computation, losing its practical benefit.
Therefore, we propose a network architecture, called separable PINN (SPINN), which can facilitate forward-mode AD for more efficient computation.
SPINN operates on a per-axis basis instead of point-wise processing in conventional PINNs, decreasing the number of network forward passes.
Besides, while the computation and memory costs of standard PINNs grow exponentially along with the grid resolution, that of our model is remarkably less susceptible, mitigating the curse of dimensionality.
We demonstrate the effectiveness of our model in various PDE systems by significantly reducing the training run-time while achieving comparable accuracy.
Project page: \url{https://jwcho5576.github.io/spinn/}\\\\
\textbf{Note}: A full paper extending this study is available at \url{https://arxiv.org/abs/2306.15969}, with additional experiments, analysis, and theorem.

\end{abstract}

\section{Introduction}
With the vast increases in computational power and breakthroughs in machine learning, researchers have explored data-driven and learning-based methods for solving partial differential equations~\cite{avrutskiy2020neural, brunton2022data, brunton2020machine, han2017deep, karniadakis2021physics, raissi2018deep, sirignano2018dgm, willard2020integrating}.
Among the promising methods, physics-informed neural networks (PINNs) have recently emerged as new data-driven PDE solvers for both forward and inverse problems~\cite{raissi2019physics}.
PINNs employ neural networks and gradient-based optimization algorithms to represent and obtain the solutions, leveraging automatic differentiation to enforce physical constraints of underlying PDEs.
It has enjoyed great success in various forward and inverse problems thanks to its numerous benefits, such as flexibility in handling a wide range of forward and inverse problems, mesh-free solutions, and not requiring observational data, hence, unsupervised training.

PINNs, using neural networks and gradient-based iterative optimization algorithms, generally require a large number of iterations to converge. 
Furthermore, the exponential growth of the number of forward and backward propagations due to the curse of dimensionality restricts their capability in solving high-dimensional PDEs. It primarily stems from using coordinate-based MLP architectures to represent the solution function, which takes input coordinates and outputs corresponding solution quantities. For each training point, computing PDE residual loss involves multiple forward and backward propagations, and the number of training points required to obtain more accurate solutions grows substantially. This paper aims to find a solution architecture to this fundamental limitation of training PINNs.

\begin{wrapfigure}{b}{0.5\textwidth}
\centering
\vspace{-1.5em}
  \includegraphics[width=0.5\textwidth]{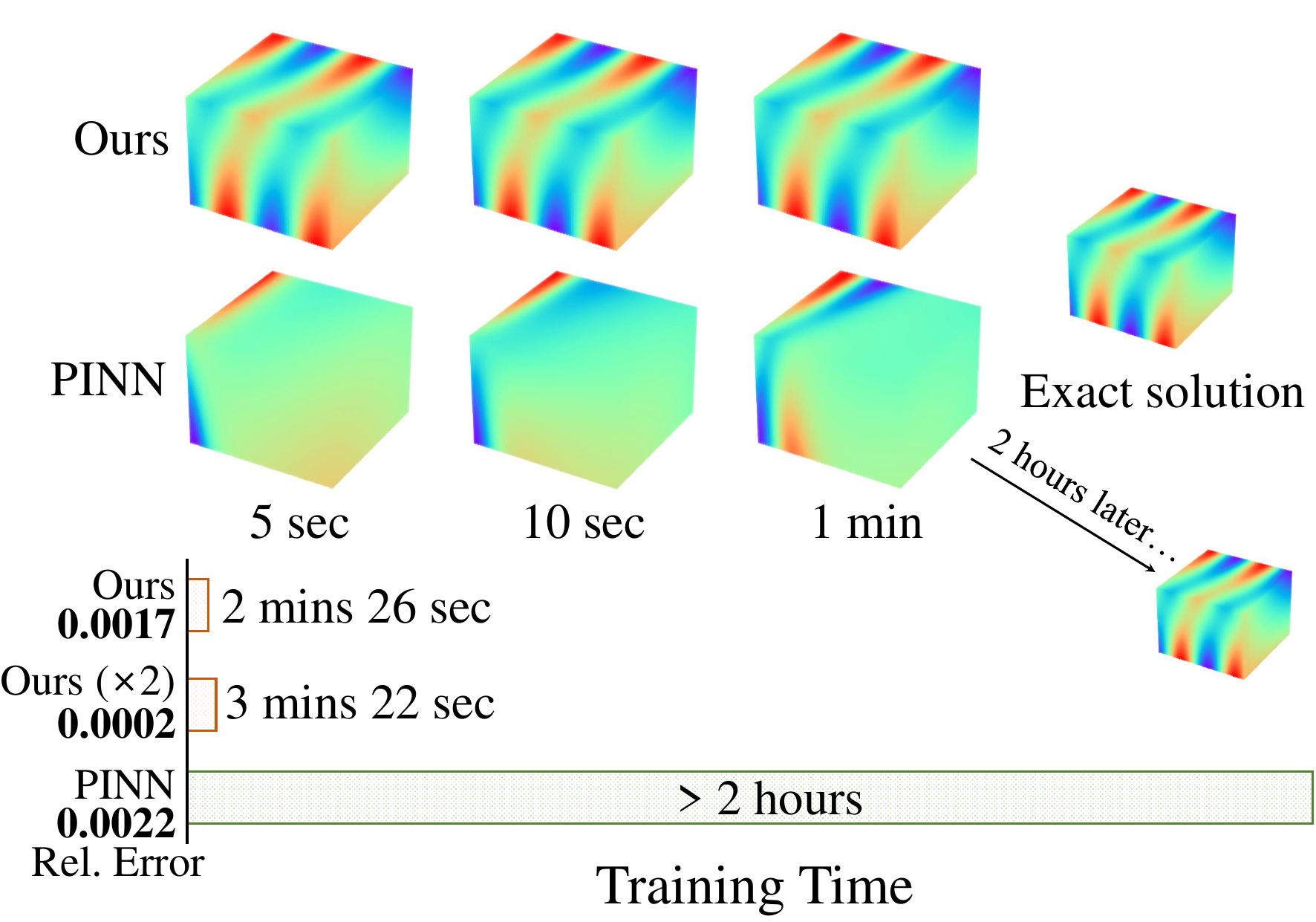}
  \caption{Training wall-clock time and relative errors compared against PINN on Klein-Gordon equation.
  `Ours ($\times2$)' denotes our model trained on $\times2$ resolution grid.
  Our model significantly reduces the computational cost without any loss in accuracy.}
  \label{fig:intro}
\vspace{-2em}
\end{wrapfigure}

We propose a novel PINN architecture, \textit{separable PINN (SPINN)}, which utilizes forward-mode automatic differentiation (AD)\footnote{a.k.a. forward accumulation mode or tangent linear mode.} to significantly reduce the computational cost for solving multi-dimensional PDEs.
Instead of feeding every multi-dimensional coordinate into a single MLP,  e.g. $f(x_1,x_2,x_3)$, we use factorized coordinates and separated sub-networks, where each sub-network takes independent one-dimensional coordinates as input, e.g. $f_1(x_1), f_2(x_2), f_3(x_3)$.
The final output is generated by a simple outer product and element-wise summation (Fig~\ref{fig:architecture}).
The suggested architecture obviates the need to query every multi-dimensional coordinate input pair, exponentially reducing the number of network propagations to generate a solution, $\mathcal{O}(n^d) \rightarrow \mathcal{O}(nd)$, where $n$ is the resolution of the solution for each dimension, and $d$ is the input dimension.

Experimental results demonstrate that given the same number of training points, SPINN significantly reduces the computational costs both in terms of FLOPs and wall-clock training time on commodity GPUs while achieving better accuracy. Figure~\ref{fig:intro} shows the run-time of training conventional PINN and our proposed method trained on a (2+1)-dimensional Klein-Gordon equation.

\section{Related Works}

\noindent \textbf{Learning-based PDE solvers}.
With the increased amount of data and powerful computing resources, data-driven and learning-based methods have shown unprecedented prosperity~\cite{avrutskiy2020neural, greenfeld2019learning, han2018solving, hennigh2021nvidia, esmaeilzadeh2020meshfreeflownet, kochkov2021machine}.
Along with massive success in the visual computing domain, early works have adopted convolutional neural networks mapping between function spaces~\cite{guo2016convolutional, esmaeilzadeh2020meshfreeflownet, zhu2018bayesian}.
In a similar vein, neural operators~\cite{anandkumar2019neural, bhattacharya2021model, li2021fourier, lu2021learning, nelsen2021random, patel2021physics} demonstrated promising results and gained attention thanks to their mesh-invariant property.
In another line of work, the unsupervised learning methods given only PDEs of dynamical systems have been suggested~\cite{bar2019unsupervised, yu2018deep, pan2020physics, raissi2019physics, smith2020eikonet}.
\\

\noindent \textbf{Physics-informed neural networks (PINNs)}.
Physics-informed neural networks (PINNs)~\cite{raissi2019physics} have received great attention as a new learning-based PDE solver.
Since its inception, many techniques have been studied to improve training PINNs for more challenging PDE systems, such as adopting curriculum learning~\cite{krishnapriyan2021characterizing}, learning-rate annealing~\cite{wang2021understanding}, causal loss function~\cite{wang2022respecting}, adaptive activation function~\cite{jagtap2020adaptive}, and adaptive coordinate sampling~\cite{daw2022rethinking}. It also has been successfully applied to real-world scientific problems, such as material sciences~\cite{chen2020physics, shukla2020physics}, plasma dynamics~\cite{mathews2020uncovering}, quantum chemistry~\cite{pfau2020ab} and medical sciences~\cite{kissas2020machine, sahli2020physics}, to name a few. 
Our method is orthogonal to the previously suggested techniques above and improves training PINNs from a computational perspective. 
\\

\begin{figure*}[t]
  \includegraphics[width=\textwidth]{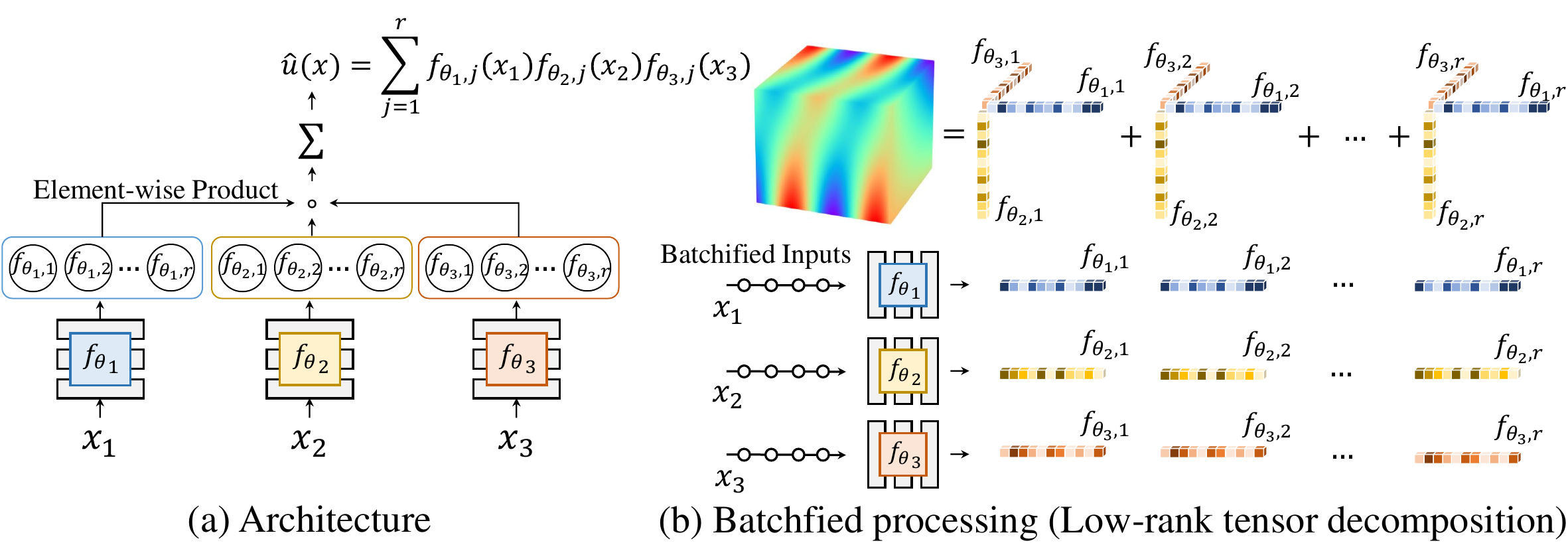}
  \vspace{-1.5em}
  \caption{(a) SPINN architecture in a 3-dimensional system.
  To solve $d$-dimensional PDE system, our model requires $d$ body MLP networks.
  (b) Batchfied processing. SPINN learns a low-rank decomposed representation of a solution.}
  \label{fig:architecture}
\end{figure*}

\noindent \textbf{Separated architectures}.
CoordX~\cite{liang2022coordx} and NAM~\cite{agarwal2021neural} have recently explored using separated MLP architectures. Unlike ours,
CoordX had to use additional layers after the feature merging step, which makes their model more computationally expensive than ours. In addition, their primary purpose is to reconstruct natural signals, such as images or 3D shapes. We focus on solving PDEs and carefully devised the architecture to exploit forward-mode differentiation to efficiently compute PDE residual losses.
NAM~\cite{agarwal2021neural} suggested separated network architectures and inputs, but only for achieving interpretability of the model's prediction.
TensoRF~\cite{chen2022tensorf} also shares similarities with ours. It used CP-decomposition to factorize the output with low-rank tensors (Figure~\ref{fig:architecture} (b)). However, TensoRF \textit{directly optimizes} low-rank tensors and uses tri-linear interpolation to implement continuous functions, or neural fields~\cite{xie2022neural}. On the other hand, we use coordinated-based MLPs to predict low-rank tensors, which is a critical difference since solving PDEs mandates computing partial derivatives with respect to input coordinates.
FBPINNs~\cite{moseley2021finite} is another approach to use multiple MLPs to solve PDEs. They decompose \textit{input domains}, and each small MLP is used to cover a particular subdomain. In contrast, we decompose \textit{input dimensions} and solve PDEs over the entire domain cooperated by all separated MLPs.
To the best of our knowledge, this is the first work to accelerate PINN training with separated MLPs leveraging forward-mode AD.
\\

\section{Forward-Mode AD with Separated Functions}

\vspace{-0.5em}

\begin{wrapfigure}{b}{0.5\textwidth}
\centering
\vspace{-1em}
  \includegraphics[width=0.5\textwidth]{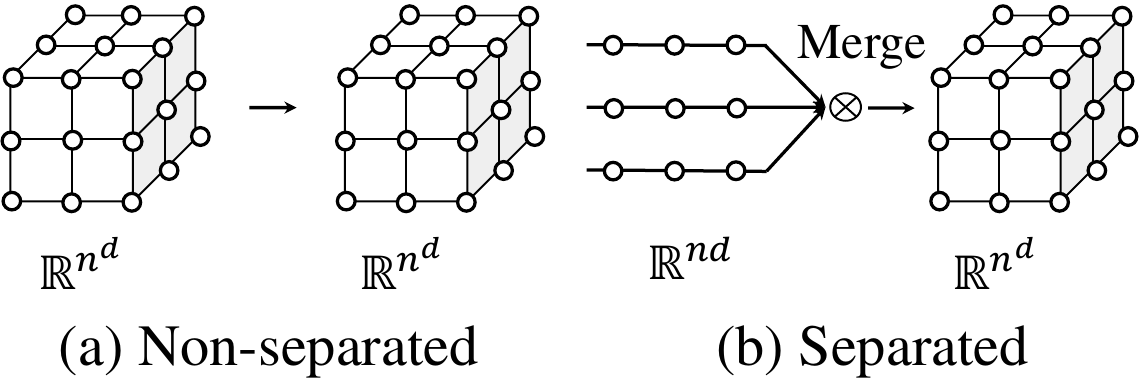}
  \caption{Non-separated vs separated. (a) $\mathbb{J}$ (Jacobian matrix) of a non-separated function approach is a $n^d \times n^d$ matrix. Computing $\mathbb{J}$ requires $\mathcal{O}(n^d)$ evaluations in both forward and reverse mode AD. (b) $\mathbb{J}$ is a $nd \times n^d$ matrix in a separated approach. $\mathcal{O}(nd)$ evaluations are required using forward-mode AD, but $\mathcal{O}(n^d)$ with reverse-mode AD.}
  \label{fig:separable_ad}
\vspace{-1em}
\end{wrapfigure}

We demonstrate that leveraging forward mode AD and separating the function into multiple functions can significantly reduce the computational cost of computing Jacobian matrix $\mathbb{J}$. In the separated approach (Fig~\ref{fig:separable_ad} (b)), we first sample $n$ one-dimensional coordinates on each of $d$ axes, which makes a total of $nd$ points.
Next, these coordinates are fed into $d$ individual functions $f_1, \hdots, f_d$ to generate the features.
We denote the number of operations of $f$ as $\mathsf{ops}(f)$.
Then, a feature merging function $g$ is used to construct the solution of the entire $n^d$ discretized domain.
The amount of computations for AD is known to be 2 or 3 times more expensive than the forward propagation~\cite{baydin2018automatic, griewank2008evaluating}. 
With the scale constants $c_f, c_g \in [2, 3]$ we can approximate the total number of operations to compute the Jacobian matrix of a separated approach,
\vspace{-0.5em}
\begin{align}
    \mathcal{C}_{\textrm{sep}}=nd c_f \mathsf{ops}(f) + c_g \mathsf{ops}(g).
    \label{eq:c_sep}
\end{align}
For a non-separated approach (Fig~\ref{fig:separable_ad} (a)),
\vspace{-0.5em}
\begin{align}
    \mathcal{C}_{\textrm{non-sep}}=n^d c_f \mathsf{ops}(f).
    \label{eq:c_non_sep}
\end{align}
If we can make $\mathsf{ops}(g)$ sufficiently small, then the ratio $\mathcal{C}_{\textrm{sep}}/\mathcal{C}_{\textrm{non-sep}}$ becomes
$\frac{nd}{n^d}\ll1$, quickly converging to 0 as $d$ and $n$ increases. 
This mitigates the curse of dimensionality for solving high-dimensional PDEs.
It also has linear complexity with respect to $n$, implying it can obtain more accurate solutions (high-resolution) efficiently.

\vspace{-1em}

\section{Separable PINN}
\label{sec:separable_pinn}
\vspace{-0.5em}
Figure~\ref{fig:architecture} (a) illustrates the overall SPINN architecture, parameterizing multiple separated functions with neural networks.
SPINN consists of $d$ body-networks (MLPs), each of which takes an individual 1-dimensional coordinate component as an input.
Each body-network $f_{\theta_i}:\mathbb{R} \rightarrow \mathbb{R}^r$ (parameterized by $\theta_i$) transforms an $i$-th input coordinate into a feature representation.
We define our predicted solution function as below:
\begin{align}
    \hat{u}(x)=\sum_{j=1}^{r}\prod_{i=1}^{d} f_{\theta_i,j}(x_i)
    \label{eq:feature_merge}
\end{align}
where $x \in \mathbb{R}^d$ is an input coordinate, $x_i$ denotes $i$-th input coordinate. $f_{\theta_i}$ is a vector-valued function and $f_{\theta_i,j}$ denotes the $j$-th component of the output vector. 
We used a tanh activation function throughout the paper.
In Eq.~\ref{eq:feature_merge}, the feature merging operation is a simple product and summation which corresponds to the merging function $g$ described in Eq.~\ref{eq:c_sep}.
Due to its simplicity, $g$ operations are cheaper than operations in MLP layers and thus, neglectable.
This makes the assumption $\mathcal{C}_{\textrm{sep}}/\mathcal{C}_{\textrm{non-sep}}\approx\frac{nd}{n^d}$ in Eq.~\ref{eq:c_sep} and~\ref{eq:c_non_sep} valid if we choose a such feature merging scheme.

The evaluated coordinate points of our model and conventional PINN have a distinct difference.
Both are \textit{uniformly} evaluated on a $d$-dimensional hypercube, but points of SPINN form a lattice-like structure, which we call as \textit{factorizable coordinates}.
This is because SPINN samples 1-dimensional coordinates from each axis and merges the features with outer product.
On the other hand, non-factorizable coordinates are random points without any structure.
Factorizable coordinates with our separated MLP architecture enable us to evaluate functions on a much more dense ($n^d$) grid, with a small number ($nd$) of inputs.
Due to its structural points and outer products between feature vectors, SPINN's solution approximation can also be viewed as a low-rank tensor decomposition (Figure~\ref{fig:architecture} (b)).
Studies of tensor approximation date back to the 1920s and among many decomposition methods, SPINN corresponds to CP-decomposition~\cite{hitchcock1927expression} which approximates a tensor by finite summation of rank-1 tensors.
While traditional methods use iterative methods such as alternating least-squares (ALS) or alternating slicewise diagonalization (ASD)~\cite{jiang2000three} to directly fit the decomposed vectors, we train neural networks which \textit{learn} a continuous representation, enabling us to take derivative w.r.t. input coordinates.


\section{Experiments}
\vspace{-0.5em}

We compared SPINN against vanilla PINN~\cite{raissi2019physics} on 3-d systems.
For every experiment, the PDE loss is calculated on fixed $90^3$ collocation points, which means we sampled $90 \cdot 3$ factorized coordinates for SPINN, while $90^3$ points are directly fed into PINN.
The number of collocation points of $90^3$ was the upper limit for the baseline model when we trained with a single NVIDIA A100 GPU with 40GB of memory.
However, the memory usage of our model was significantly smaller, enabling SPINN to use more collocation points ($180^3$ and $360^3$) to get more accurate solutions.
All experiments are trained with full batch for both models and reported error metrics are average relative $L_2$ errors computed by $\lVert\hat{u}-u\rVert^2/\lVert u\rVert^2$, where $\hat{u}$ is the model prediction and $u$ is the reference solution.
All experiments were performed five times with different random seeds.

\begin{figure}[ht]
    \begin{minipage}{.45\linewidth}
      \centering
      \scriptsize{
        \begin{tabular}{@{\hskip 0.3em}c@{\hskip 0.3em}|@{\hskip 0.3em}c@{\hskip 0.3em}|@{\hskip 0.3em}c@{\hskip 0.3em}c@{\hskip 0.3em}c@{\hskip 0.3em}}
            \toprule
            \multirow{2}{*}{system (3-d)} & \multirow{2}{*}{model} & relative & memory & run-time \\
            & & $L_2$ error & (MB) & (ms/iter.) \\
            \bottomrule
            & PINN ($90^3$) & 0.0045$\pm$0.0002 & 35,640 & 129.84 \\ 
            linear & Ours ($90^3$) & 0.0031$\pm$0.0002 & 832 & 2.64 \\ 
            diffusion & Ours ($180^3$) & \textbf{0.0029}$\pm$0.0002 & 964 & 3.59 \\
            & Ours ($360^3$) & 0.0031$\pm$0.0002 & 1,596 & 7.48 \\
            \midrule
            & PINN ($90^3$) & 0.0076$\pm$0.0003 & 35,640 & 139.82 \\ 
            nonlinear & Ours ($90^3$) & 0.0061$\pm$0.0009 & 864 & 3.01 \\ 
            diffusion & Ours ($180^3$) & \textbf{0.0047}$\pm$0.0006 & 1,348 & 4.51 \\
            & Ours ($360^3$) & 0.0049$\pm$0.0007 & 5,180 & 11.59 \\
            \midrule
            \multirow{4}{*}{Helmholtz} & PINN ($90^3$) & 0.1501$\pm$0.0272 & 35,640 & 166.06 \\ 
            & Ours ($90^3$) & 0.0730$\pm$0.0369 & 856 & 2.95 \\ 
            & Ours ($180^3$) & 0.0623$\pm$0.0143 & 964 & 4.16 \\
            & Ours ($360^3$) & \textbf{0.0514}$\pm$0.0204 & 2,108 & 9.57 \\
            \midrule
            \multirow{4}{*}{Klein-Gordon} & PINN ($90^3$) & 0.0092$\pm$0.0022 & 35,640 & 166.44 \\ 
            & Ours ($90^3$) & 0.0016$\pm$0.0017 & 888 & 2.93 \\ 
            & Ours ($180^3$) & \textbf{0.0009}$\pm$0.0002 & 1,132 & 4.04 \\
            & Ours ($360^3$) & \textbf{0.0009}$\pm$0.0003 & 3,164 & 9.13 \\
            \bottomrule
        \end{tabular}
        }
    \captionof{table}{Overall results of 3-d systems.
    $n^3$ denotes the number of collocation points.
    The training run-time (milliseconds per iteration) of our model is substantially lower than the baseline while achieving higher accuracy.}
    \label{table:3d_result}
    \end{minipage}%
    \begin{minipage}{.5\linewidth\;\;\;}
    \includegraphics[width=\textwidth]{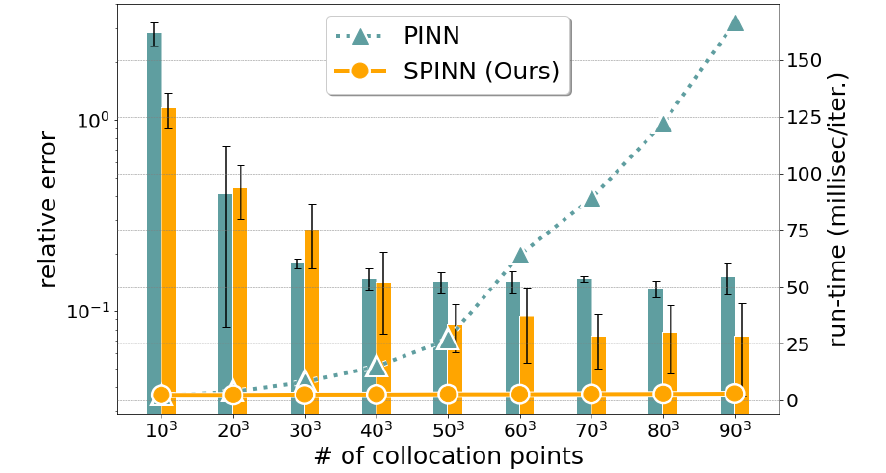}
    \caption{Our model mitigates the curse of dimensionality.
    It shows the relative error and run-time of the Helmholtz equation, on a varying number of collocation points.
    The bar graph shows the relative $L_2$ error (left $y$-axis), and the line graph shows the run-time in milliseconds (right $y$-axis) per iteration.
    We get more accurate solutions as the number of collocation points increases.
    The run-time of the baseline model grows exponentially with respect to the resolution, while our model keeps it almost constant.}
    \label{fig:helmholtz_graph}
    \end{minipage} 
\end{figure}

Table~\ref{table:3d_result} shows our overall results. SPINN is significantly more computationally efficient than the baseline PINN in terms of wall-clock run-time.
Klein-Gordon equation showed the largest gap, where our model achieved $57\times$ speed-up given the same input size.
With the help of the method outlined in~\cite{griewank2008evaluating}, we also estimated the FLOPs for evaluating the derivatives.
In comparison to the baseline, SPINN requires $1,195\times$ less computations to evaluate the forward pass, first and second order derivatives.
See section~\ref{sec:flops} in the appendix for details.

\begin{figure*}[ht]
  \includegraphics[width=\textwidth]{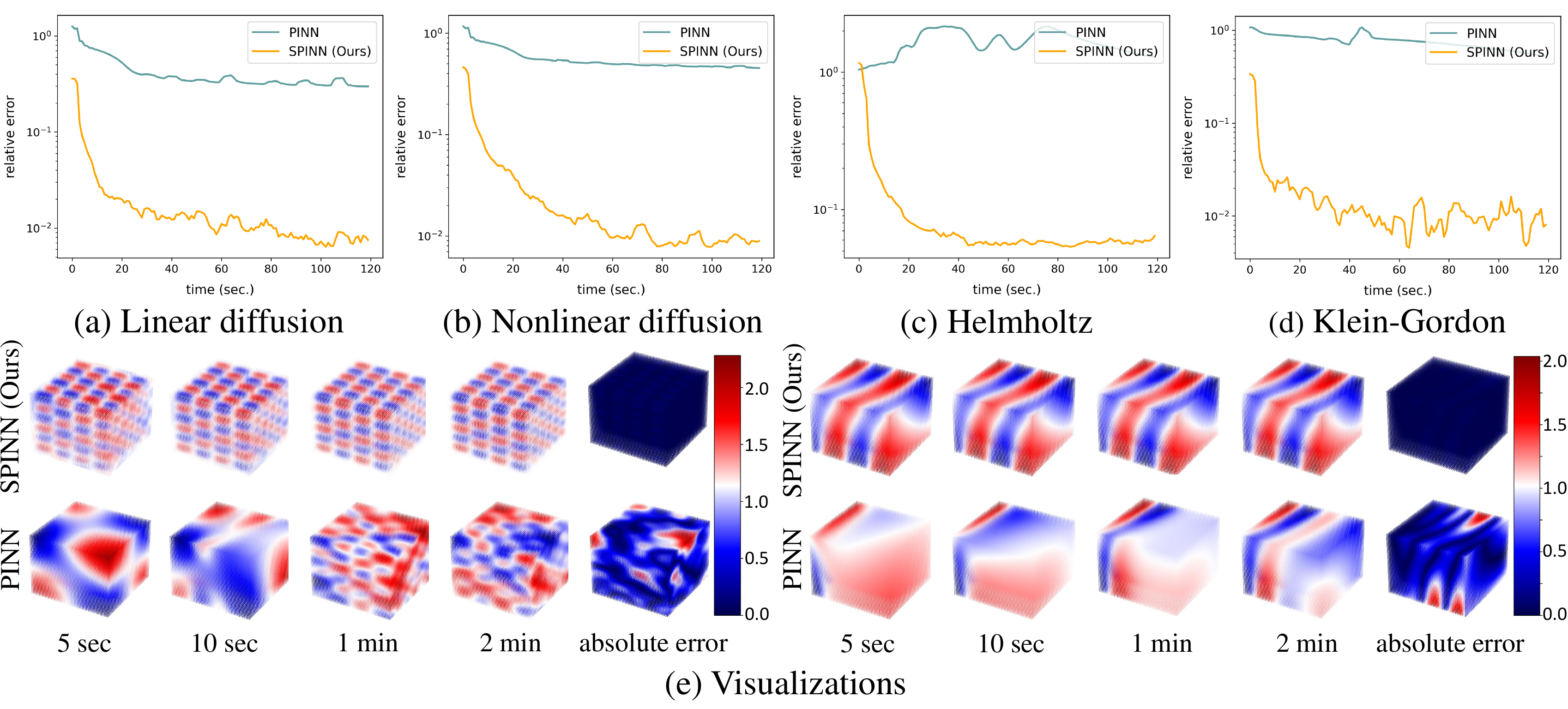}
  \caption{(a) to (d) Error curves of each experiment trained on $90^3$ collocation points.
  The $x$-axis shows the actual run-time in seconds during training.
  (e) Visualized solutions at 4 time stamps (left: 3-d Helmholtz, right: 3-d Klein-Gordon). Absolute error is plotted at training time 2 min.}
  \label{fig:visualization}
\end{figure*}

In all cases, our model finds the most accurate solution.
Furthermore, while the number of collocation points grows exponentially, the memory usage and actual run-time of our model show logarithmic growth (Figure~\ref{fig:helmholtz_graph}).
SPINN is also memory-efficient because it stores a much smaller batch of tensors, which are the intermediate values (primals) while building the computational graph.
We also identified from the results that we get more accurate solutions with more collocation points.
This training characteristic of PINN substantiates that our method is very effective for solving multi-dimensional PDEs.
Figure~\ref{fig:visualization} shows the error curves with run-time and visualized solutions of Helmholtz and Klein-Gordon equation.
Additional experimental details and analyses are provided in the appendix.

\noindent \textbf{Acknowledgements}
This research was supported by the Ministry of Science and ICT (MSIT) of Korea, under the National Research Foundation (NRF) grant (2022R1F1A1064184, 2022R1A4A3033571), Institute of Information and Communication Technology Planning Evaluation (IITP) grants for the AI Graduate School program (IITP-2019-0-00421). The research of Seok-Bae Yun was supported by Samsung Science and Technology Foundation under Project Number SSTF-BA1801-02.

\bibliographystyle{plain}
\bibliography{reference}

\clearpage

\appendix

\section{Forward/Reverse-Mode AD}

\begin{wraptable}{r}{8.5cm}
\vspace{-0.5em}
\scriptsize
\setlength\tabcolsep{0.5pt}
\centering
    \begin{tabular*}{0.6\columnwidth}{ll ll lll}
        \toprule
        \multicolumn{2}{l}{Forward primal trace} & \multicolumn{2}{c}{Forward tangent trace} & \multicolumn{3}{l}{Backward adjoint trace} \\
        \tikzmark{a} & $v_{0} = x$ & \tikzmark{c} & $\dot{v}_{0} = \dot{x}$ & \tikzmark{e} & \multicolumn{2}{l}{$\bar{x}\mspace{6mu} = \bar{v}_{0}$} \\ \cline{2-2} \cline{4-4} \cline{6-7}
        & $v_1 = W_1 \cdot v_0$ && $\dot{v}_{1} = W_1 \cdot \dot{v}_{0}$ &&
        $\bar{v}_{0} = \bar{v}_1 \cdot \frac{\partial{v_1}}{\partial{v}_{0}}$ & $=\bar{v}_{1} \cdot W_1$ \\
        & $v_2 = \text{tanh}(v_1)$ && $\dot{v}_2 = \text{tanh}^{'}(v_1) \circ \dot{v}_{1}$ &&
        $\bar{v}_{1} = \bar{v}_{2} \cdot \frac{\partial{v_2}}{\partial{v_1}}$ & $=\bar{v}_{2} \circ \text{tanh}^{'}(v_1)$ \\
        & $v_3 = W_2 \cdot v_2$ && $\dot{v}_{3} = W_2 \cdot \dot{v}_{2}$ &&
        $\bar{v}_{2} = \bar{v}_{3} \cdot \frac{\partial{v_3}}{\partial{v_2}}$ & $=\bar{v}_{3} \cdot W_2$ \\
        \cline{2-2} \cline{4-4} \cline{6-7}
        \tikzmark{b} & $y\mspace{6mu} = v_3$ & \tikzmark{d} & $\dot{y}\mspace{6mu} = \dot{v}_3$ & \tikzmark{f} & \multicolumn{2}{l}{$\bar{v}_{3} = \bar{y}$} \\
        \bottomrule
    \end{tabular*}
    \tikz[remember picture,overlay] \draw[-latex, line width=0.25mm] (a.north -| b.south) -- (b.south);
    \tikz[remember picture,overlay] \draw[-latex, line width=0.25mm] (c.north -| d.south) -- (d.south);
    \tikz[remember picture,overlay] \draw[-latex, line width=0.25mm] (f.south -| e.north) -- (e.north);
    \caption{An example of forward and reverse mode AD in a two-layers tanh MLP. Here $v_0$ denotes the input variable, $v_k$ the primals, $\dot{v}_k$ the tangents, $\bar{v}_k$ the adjoints, and $W_1, W_2$ the weight matrices. Biases are omitted for brevity.}
    \label{table:ad_trace}
\end{wraptable}

In this section, we provide some background on the two modes of AD and how Jacobian matrices are evaluated.
For clarity, we will follow the notations used in~\citenew{baydin2018automaticsupp} and~\citenew{griewank2008evaluatingsupp}.
Suppose our function $f:\mathbb{R}^n\rightarrow\mathbb{R}^m$ is a two-layers MLP with tanh activation.
First column of Table~\ref{table:ad_trace} demonstrates a single forward trace of $f$.
To obtain a $m \times n$ Jacobian matrix $\mathbb{J}_f$ in forward-mode, we compute the Jacobian-vector product (JVP),
\begin{align}
    \mathbb{J}_{f} r = 
    \begin{bmatrix}
        \frac{\partial y_1}{\partial x_1} & \hdots & \frac{\partial y_1}{\partial x_n}\\
        \vdots & \ddots & \vdots\\
        \frac{\partial y_m}{\partial x_1} & \hdots & \frac{\partial y_m}{\partial x_n}
    \end{bmatrix}
    \begin{bmatrix}
        \partial x_1/\partial x_i\\
        \vdots\\
        \partial x_n/\partial x_i
    \end{bmatrix},
\end{align}
for $i\in\{1,\hdots,n\}$.
The forward-mode AD is a one-phase process: while tracing primals (intermediate values), $v_k$, it continues to evaluate and accumulate their tangents $\dot{v_k}=\partial v_k/\partial x_i$.
This is equivalent to decomposing one large JVP into a series of JVPs by the chain rule and computing them from right to left.
A run of JVP with the initial tangents $\dot{v_0}$ as the first column vector of an identity matrix $I_n$ gives the first column of $\mathbb{J}_f$.
Thus, the full Jacobian can be obtained in $n$ evaluations.

On the other hand, the reverse-mode AD computes vector-Jacobian product (VJP):
\begin{align}
    r^\top \mathbb{J}_{f} = 
    \begin{bmatrix}
        \partial y_j/\partial y_1 & \hdots & \partial y_j/\partial y_m\\
    \end{bmatrix}
    \begin{bmatrix}
        \frac{\partial y_1}{\partial x_1} & \hdots & \frac{\partial y_1}{\partial x_n}\\
        \vdots & \ddots & \vdots\\
        \frac{\partial y_m}{\partial x_1} & \hdots & \frac{\partial y_m}{\partial x_n}
    \end{bmatrix},
\end{align}
for $j\in\{1,\hdots,m\}$, which is the reverse-order operation of JVP.
This is a two-phase process.
The first phase corresponds to forward propagation, storing all the primals, $v_k$, and recording the elementary operations in the computational graph.
In the second phase, the derivatives are computed by accumulating the adjoints $\bar{v}_k=\partial y_j/\partial v_k$ (third of Table~\ref{table:ad_trace}).
Since VJP builds one row of a Jacobian at a time, it takes $m$ evaluations to obtain the full Jacobian.
To sum up, the forward-mode is more efficient for a tall Jacobian ($m>n$), while the reverse-mode is better suited for a wide Jacobian ($n>m$).
Please refer to~\citenew{baydin2018automaticsupp} for a more comprehensive review.



\section{Training with Physics-Informed Loss}
\label{sec:pinn_loss}

After SPINN predicts an output function with the methods described in section~\ref{sec:separable_pinn}, the rest of the training procedure follows the same process used in conventional PINN training~\citenew{raissi2019physicssupp}, except we use forward mode AD (standard back-propagation, a.k.a. reverse mode AD, used for parameter updates). With the slight abuse of notation, our predicted solution function is denoted as $\hat{u}(x,t)$, explicitly expressing time coordinates.
Given an underlying PDE (or ODE) and initial, boundary conditions, SPINN is trained with a ``physics-informed'' loss function:
\begin{gather}
    \min_{\theta}\mathcal{L}(\hat{u}_\theta(x,t))=\min_\theta\lambda_{\textrm{pde}}\mathcal{L}_{\textrm{pde}}+\lambda_{\textrm{ic}}\mathcal{L}_{\textrm{ic}}+\lambda_{\textrm{bc}}\mathcal{L}_{\textrm{bc}},
    \label{eq:objective_function}\\
    \mathcal{L}_{\textrm{pde}}=\int_\Gamma\int_\Omega\lVert\mathcal{N}_{x,t}[\hat{u}_\theta](x,t)\rVert^2dxdt,
    \label{eq:pde_loss}\\
    \mathcal{L}_{\textrm{ic}}=\int_\Omega\lVert \hat{u}_\theta(x,0)-u_{\textrm{ic}}(x)\rVert^2dx,\\
    \mathcal{L}_{\textrm{bc}}=\int_\Gamma\int_{\partial\Omega}\lVert\mathcal{B}_{x,t}[\hat{u}_\theta](x,t)-u_{\textrm{bc}}(x,t)\rVert^2dxdt,
\end{gather}
where $\Omega$ is an input domain, $\mathcal{N}, \mathcal{B}$ are generic differential operators and $u_{\textrm{ic}}, u_{\textrm{bc}}$ are initial, boundary conditions, respectively.
$\lambda$ are weighting factors for each loss term.
When calculating the PDE loss ($\mathcal{L}_{pde}$) we sampled collocation points from factorized coordinates and used forward-mode AD.
The remaining $\mathcal{L}_{\textrm{ic}}$ and $\mathcal{L}_{\textrm{bc}}$ are then computed with initial and boundary coordinates to regress the given conditions.
By minimizing the objective loss in Eq.~\ref{eq:objective_function}, the model output is enforced to satisfy the given equation and the initial, boundary conditions.

\section{FLOPs Estimation}
\label{sec:flops}
The FLOPs for evaluating the derivatives can be systematically calculated by disassembling the computational graph into elementary operations such as additions and multiplications.
Given the computational graph of forward pass for computing the primals, AD augments each elementary operation into other elementary operations.
The FLOPs in the forward pass can be calculated precisely since it consists of a series of matrix multiplications and additions.
We used the method described in~\citenew{griewank2008evaluatingsupp} to estimate FLOPs for evaluating the derivatives.
Table~\ref{tab:flops} shows the number of additions (\textsf{ADDS}), and multiplications (\textsf{MULTS}) in each evaluation process.
Note that FLOPs is a summation of \textsf{ADDS} and \textsf{MULTS} by definition.

\begin{table}[!htb]
\centering
    \begin{tabular}{c|c|c|c|c}
        \toprule
         & \multicolumn{2}{c|}{SPINN (ours)} & \multicolumn{2}{c}{PINN (baseline)}\\
        \midrule
         & \textsf{ADDS} ($\times10^6$) & \textsf{MULTS} ($\times10^6$) & \textsf{ADDS} ($\times10^6$) & \textsf{MULTS} ($\times10^6$) \\ 
        \midrule
        forward pass & 39 & 40 & 36,742 & 36,742 \\
        1st-order derivative & 79 & 80 & 147,404 & 73,921 \\
        2nd-order derivative & 159 & 160 & 221,324 & 148,279 \\
        \midrule
        MFLOPs (total) & \multicolumn{2}{c|}{556} & \multicolumn{2}{c}{664,411} \\
        \bottomrule
    \end{tabular}
    \caption{The number of elementary operations for evaluating forward pass, first and second-order derivatives.
    The calculation is based on $90^3$ collocation points in a 3-d system and the MLP settings described in section~\ref{sec:exp_detail} for both models.
    We assumed that each derivative is evaluated on every coordinate axis.
    }
    \label{tab:flops}
\end{table}

\section{Experimental Details and Analyses}
\label{sec:exp_detail}

We matched the total number of learnable parameters between each model for a fair comparison.
For our model, we used 3 body networks of 5 hidden layers with 50 hidden/output feature sizes each.
For the baseline model, we used a single MLP of 5 hidden layers with 100 hidden feature sizes.
We used Adam~\citenew{kingma2015adamsupp} optimizer with a learning rate of 0.001 and trained for 50,000 iterations for every experiment.
All weight factors $\lambda$ in the loss function in Eq.~\ref{eq:objective_function} are set to 1. 
In the following sections, we give more detailed descriptions and analyses of each experiment.

\subsection{Linear/Non-linear Diffusion Equation}
Diffusion equation is one of the most representative parabolic PDEs, often used for modeling the heat diffusion process. It can be written as:
\begin{align}
    &\partial_tu = \alpha\Delta u, &x\in\Omega, t\in\Gamma,\label{eq:linear_diffusion}\\
    &\partial_tu = \alpha\left(\lVert\nabla u\rVert^2 + u\Delta u\right), &x\in\Omega, t\in\Gamma,\label{eq:nonlinear_diffusion}\\
    &u(x, 0) = u_{ic}(x), &x\in\Omega,\\
    &u(x, t) = 0, &x\in\partial\Omega, t\in\Gamma,
\end{align}
where Eq.~\ref{eq:linear_diffusion} is a linear form and Eq.~\ref{eq:nonlinear_diffusion} is a non-linear form.
We used diffusivity $\alpha=0.05$, spatial domain $\Omega\in[-1,1]^2$, temporal domain $\Gamma\in[0,1]$ and used superposition of three Gaussian functions for initial condition $u_{\textrm{ic}}$.
For error measurement, we compared against the reference solution obtained by a widely-used PDE solver platform FEniCS~\citenew{logg2012automatedsupp}.
Note that FEniCS is a FEM-based solver.
The first and second row of Table~\ref{table:3d_result} shows the numerical results of the linear and non-linear diffusion equation, respectively.
With $40\times$ less memory usage, our method finds more accurate solutions, $49\times$ ($47\times$ for nonlinear case) faster than the baseline where $180^3$ collocation points showed the lowest error.

\subsection{Helmholtz Equation}
The Helmholtz equation is a time-independent wave equation that takes the form:
\begin{align}
    &\Delta u +k^2u = q, &x\in\Omega,\\
    &u(x) = 0, &x\in\partial\Omega,
\end{align}
where the spatial domain is $\Omega\in[-1,1]^3$.
For a given source term $q=-(a_1\pi)^2u-(a_2\pi)^2u-(a_3\pi)^2u+k^2u$, we devised a manufactured solution $u=\sin(a_1\pi x_1)\sin(a_2\pi x_2)\sin(a_3\pi x_3)$, where we take $k=1, a_1=3, a_2=3, a_3=2$.
The results of our experiments are shown in the third row of Table~\ref{table:3d_result}.
Due to stiffness in the gradient flow, it hinders conventional PINNs to find accurate solutions.
Therefore, a learning rate annealing algorithm is suggested to mitigate such phenomenon~\citenew{wang2021understandingsupp}.
However, without bells and whistles, SPINN successfully finds accurate solutions.
On the same number of collocation points, the memory usage of our model is $41\times$ smaller, and the training speed is $56\times$ faster than the baseline.
Visualized solutions are illustrated in Figure~\ref{fig:visualization} (e) left.

Helmholtz equation contains three second-order derivative terms, which makes the baseline model most time-consuming (more than two hours for full training).
Meanwhile, the linear diffusion equation contains one first-order derivative and two second-order derivative terms, which has fewer second-order derivatives than Helmholtz.
Comparing Helmholtz's experiment against the linear diffusion experiment, the training time increased 28\% ($129.84\rightarrow166.06$) for the baseline model, but SPINN has only increased 12\% ($2.64\rightarrow2.95$).
This is due to SPINN's significant reduction of the forward-mode AD computations, which makes our model also very efficient at solving high-order PDEs.

Figure~\ref{fig:helmholtz_graph} also shows the result of training Helmholtz equation to identify the error and run-time dynamics along the number of collocation points.
We increased the resolution of each axis from 10 to 90 (maximum size for baseline) and recorded the relative errors and training run-times for each resolution.
As expected, the relative errors of both models tend to decrease along the resolution.
However, the errors of the baseline model starts to converge at about $40^3$, while our model keeps finding a more accurate solution with larger points.
More importantly, the training run-times of our model barely increase ($2.43\rightarrow2.95$) while that of the baseline model is very sensitive to the resolution ($2.13\rightarrow166.06$).

\subsection{Klein-Gordon Equation}
The Klein-Gordon equation is a non-linear hyperbolic PDE, which arises in diverse applied physics for modeling relativistic wave propagation.
The inhomogeneous Klein-Gordon equation is given by
\begin{align}
    &\partial_{tt}u - \Delta u + u^2 = f, &x\in\Omega, t\in\Gamma \\
    &u(x,0)=x_1+x_2, &x\in\Omega \\
    &u(x,t)=u_{bc}(x), &x\in\partial\Omega, t\in\Gamma,
\end{align}
where we chose the spatial/temporal domain to be $\Omega\in[-1,1]^2$ and $\Gamma\in[0,10]$, respectively.
For error measurement, We used a manufactured solution $u=(x_1+x_2)\cos(t)+x_1x_2\sin(t)$ and $f$, $u_{bc}$ are extracted from this exact solution.
Fourth row of Table~\ref{table:3d_result} shows the quantitative results.
Again, our method finds the best solution in every resolution with much efficient memory usage ($40\times$) and training run-time ($57\times$).
Note that this equation showed the largest difference in terms of run-time.
Analogous to Helmholtz equation, Klein-Gordon equation contains the same number of second-order derivatives and thus, shows a similar aspect on training run-time.
Visualized results are shown in Figure~\ref{fig:visualization} (e) right.

\bibliographystylenew{plain}
\bibliographynew{supple}

\end{document}